\documentclass[a4paper,conference]{IEEEtran}
\ifCLASSINFOpdf
   \usepackage[pdftex]{graphicx}
  \graphicspath{ {./figs/} }  
  \usepackage{subfig}  
\else
\fi
\usepackage{amsmath}
\usepackage{amssymb}  
\usepackage{gensymb}  
\usepackage{multirow}  
\usepackage{tabularx}  

\begin{document}
%
\title{GazeMAE: General Representations of Eye Movements using a Micro-Macro Autoencoder}

\author{\IEEEauthorblockN{Louise Gillian C. Bautista}
\IEEEauthorblockA{Department of Computer Science\\
University of the Philippines\\
Email: lcbautista1@up.edu.ph}
\and
\IEEEauthorblockN{Prospero C. Naval, Jr.}
\IEEEauthorblockA{Department of Computer Science\\
University of the Philippines\\
Email: pcnaval@up.edu.ph}
}


%


\maketitle

\begin{abstract}

Eye movements are intricate and dynamic events that contain a wealth of information about the subject and the stimuli. We propose an abstract representation of eye movements that preserve the important nuances in gaze behavior while being stimuli-agnostic. We consider eye movements as raw position and velocity signals and train separate deep temporal convolutional autoencoders. The autoencoders learn micro-scale and macro-scale representations that correspond to the fast and slow features of eye movements.
We evaluate the joint representations with a linear classifier fitted on various classification tasks. Our work accurately discriminates between gender and age groups and outperforms previous works on biometrics and stimuli classification.
Further experiments highlight the validity and generalizability of this method, bringing eye-tracking research closer to real-world applications.
\end{abstract}


%
\IEEEpeerreviewmaketitle

\section{Introduction}

Our eyes move in response to top-down and bottom-up factors, subconsciously influenced by a stimuli's characteristics and our own goals \cite{cognition-konig-2016}. Eye movements can be seen simply as a sequence of fixations and saccades: at some points we keep our eyes still to take in information, then rapidly move them to switch our point of focus. Thus, eye movements tell a lot about our perception, thought, and decision-making processes \cite{saccades-hutton-2008}.  In addition, there exist less-pronounced eye movements even within a fixation, among them are microsaccades that have recently been found to have numerous links to attention, memory, and cognitive load \cite{etra, siegenthaler-2013, fem-martinezconde-2017, krejtz-2018}. Overall, such findings encourage eye-tracking technology to be brought to various fields such as human-computer interaction, psychology, education, medicine, and security \cite{duchowski}.

Bridging the gap between laboratory findings and real-world applications require that eye movements are processed using representations or feature vectors as inputs to algorithms. Common methods to do so include processing gaze into parameters \cite{features-rigas-2018} (e.g. fixation counts and durations), maps \cite{lemeur2012-scanpaths} (e.g. heat maps, saliency maps), scanpaths \cite{anderson2014-scanpaths} (e.g. string sequences), and graphical models \cite{gant-cantoni-2015, coutrot2017-hmm} that consider image regions as nodes and saccades as edges.

However, these methods have two main drawbacks that inhibit them from optimally representing eye movements. First, they do not exploit the wealth of information present in eye movements. By discretizing movements into fixations and saccades, they flatten the dynamic nature of eye movements and lose the tiny but important nuances. Additionally, event detection is still an active research area and as such may be prone to inaccuracies and inconsistencies \cite{eye-tracking-andersson-2017, hessels-2018}. Second, they are not generalizable due to the tight links of the methods to the stimuli, thereby limiting eye movement comparison to those elicited from the same image or stimuli. Scanpaths and graphs additionally have a dependence on pre-defined areas of interest (AoIs). This may be mitigated by learning AoIs in a data-driven manner, but this in turn introduces dependencies on the method and on the amount and quality of data available for each new stimulus.

\begin{figure}[tbh]
    \centering
    \includegraphics[width=\columnwidth]{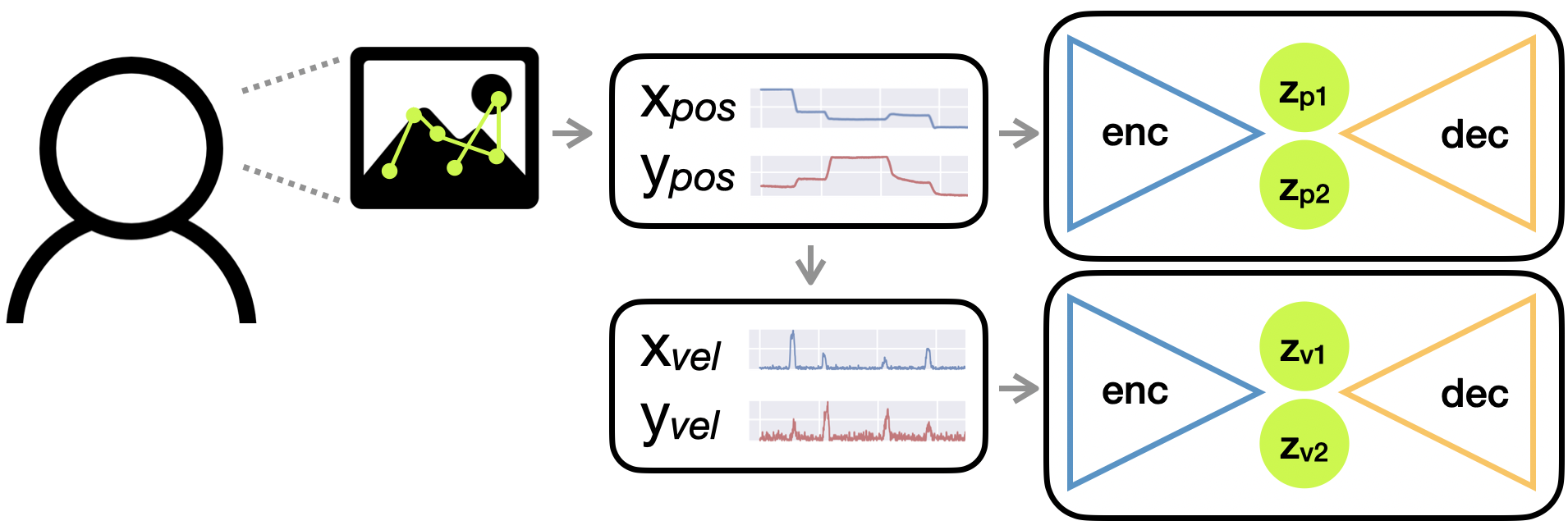}
    \caption{Raw eye movement position and velocity signals are used as input to autoencoders which learn micro-scale ($z_{p1}$, $z_{v1}$) and macro-scale ($z_{p2}$, $z_{v2}$) representations.}
    \label{fig:overview}
\end{figure}

In this work, we use deep unsupervised learning to learn abstract representations of eye movements. This removes the need for extensive feature engineering, allowing us to bypass the event detection steps and learn from the full resolution of the data. We use only the position and velocity signals as input, making this method stimuli-agnostic. It can extract representations for any sample regardless of stimuli, enabling comparisons to be made.
In particular, we use an autoencoder (AE) in which the encoder and decoder networks are temporal convolutional networks (TCN). Our AE architecture uses two bottlenecks, encoding information at a micro and macro scale. We train a model on position signals, and another on velocity signals. The models are evaluated on various classification tasks with a linear classifier.
Characteristics such as identity, age, gender, and stimuli were predicted using AE representations. Additionally, we show that the AE can handle any input length (i.e. viewing time), generalize to an unseen data set with a lower sampling frequency, and perform comparably with a supervised version of the encoder network.
\newline

The contributions of this paper are as follows:
\begin{enumerate}
    \item We apply deep unsupervised learning to eye movement signals such that representations are learned without supervision or feature engineering.
    \item We learn representations for eye movements that are stimuli-agnostic.
    \item We propose a modified autoencoder with two bottlenecks that learn fast and slow features of the eye movement signal. This autoencoder also uses an interpolative decoder instead of a regular Temporal Convolutional Network or an autoregressive decoder.
    \item We show that the representations learned are meaningful. They are able to accurately classify labels, generalize to an unseen data set, scale to long input lengths. Furthermore, similar data points exhibit clustering properties.
\end{enumerate}
Note that this work is limited to eye movements gathered on static and visual stimuli, recorded with research-grade eye-trackers. Eye movements on texts, videos, or "in the wild" are beyond our scope. Source code and models are available at https://github.com/chipbautista/gazemae.
\section{Methodology}

\subsection{Preliminaries}
\subsubsection{Representation Learning}
The goal of representation learning, also called feature learning, is to abstract information from data such that the underlying factors of variation in the data are captured \cite{rep-learning-bengio-2013}. This involves mapping an input to an embedding space which meaningfully describe the original data. A common use case for learning representations is to act as a preprocessing step for downstream tasks in which the representation, often notated as $z$ of a data point $x$, will be used as the input for classifiers and predictors. Representation learning methods are commonly unsupervised methods, where no external labels about the data is required. Therefore, these can take advantage of any available data to learn more robust features.

\subsubsection{Autoencoder}
An autoencoder (AE) is a neural network that learns a representation of an input data by attempting to reconstruct a close approximation of the input. A typical AE is undercomplete, i.e. it uses a bottleneck to compresses the input to a lower-dimensional space before producing an output with the same dimensions as the input.

Generally, an AE works as follows: an encoder $f(x)$ maps the original input $x \in \mathbb{R}^{d_x}$ to a latent vector $z \in \mathbb{R}^{d_z}$, and a decoder $g(z)$ maps $z$ to an output $\hat{x} \in \mathbb{R}^{d_x}$.
It is trained to reconstruct $x$, i.e. $\hat{x} \approx x$.

Since $d_z < d_x$, the encoder is forced to learn only the relevant information such that the decoder $g$ is able to sufficiently reconstruct the original input. This is a simple framework to learn a representation of the data, and is commonly thought of as a non-linear version of Principal Component Analysis (PCA) \cite{rep-learning-bengio-2013}. Because an AE uses the input data as its target output, it is a self-supervised method for representation learning.

\subsubsection{Temporal Convolutional Network}
The temporal convolutional network (TCN) \cite{tcn} is a generic convolutional neural network (CNN) architecture that has recently been shown to outperform Recurrent Neural Networks (RNNs). TCNs work in the same manner as the CNN, where each convolutional layer convolves a number of 1-dimensional kernels ($c$ filters of size $1 \times k$) across the input data to recognize sequence patterns \cite{convolutions}. A TCN modifies the convolution operation into the following:
\begin{enumerate}
    \item \emph{Dilated} Convolutions, where the kernel skips $d-1$ values. For a learnable kernel $K$ with kernel size $k$ and dilation $d$, the output at a subsequence $x$ of size $n$ in an input $X$ is calculated with the following:
    \begin{equation*}
        \sum_{i=0}^{k} K_i \cdot x_{n-di}
    \end{equation*}
     Dilations are commonly increased exponentially across layers, e.g. $2^0, 2^1, 2^2, ... 2^l$. This enables the output in layer $l$ to be calculated with higher receptive field i.e. from a wider input range.
    \item \emph{Causal} Convolutions The output at time $t$ is calculated using only the values from the previous time steps $t-1, t-2, t-3, ...$. This is done by padding $d(k - 1)$ zeroes on the left of the input. In effect, this emulates the sequential processing of RNNs.
\end{enumerate}

\subsection{Data Sets}

\subsubsection{EMVIC}
The Eye Movements Verification and Identification Competition (EMVIC) 2014 \cite{emvic} is a data set used as a benchmark for Biometrics, where subjects are to be identified based only on their eye movements. They collected data from 34 subjects who were shown a number of normalized face images (the eyes, nose, and mouth are in roughly the same position in the images).
The viewing times spent by the subjects to look at the face images range from 891 ms to 22012 ms, and the average is 2429 ms or roughly 2.5 seconds. Eye movements were recorded using a Jazz-Novo eye tracker with a 1000 Hz sampling frequency, i.e. it records 1000 gaze points per second. 1,430 eye movement samples were collected, where the training set consists of 837 samples from 34 subjects and the test set consists of 593 from 22 subjects. 

\subsubsection{FIFA}
The Fixations in Faces (FIFA) \cite{fifa} is an eye movement data set of 7 subjects using 250 images from indoor and outdoor scenes.
Eye movements were recorded using SR Research EyeLink 1000 eye-tracker with a 1000 Hz sampling frequency. The images were of 1024x768 resolution and were displayed on a screen 80cm from the subject. This corresponds to a subjects' visual angle of 28\degree x 21\degree. We obtain 3,200 samples from this data set.

\subsubsection{ETRA}

The Eye Tracking Research \& Applications (ETRA) data set was used to analyze saccades and microsaccades in \cite{etra, etra-2} and was also used for a data mining challenge in ETRA 2019.
Eight subjects participated and viewed 4 image types: blank image, natural scenes, picture puzzles, and "Where's Waldo?" images. 
For the blank and natural image types, the subjects were free to view the image in any manner. Picture puzzles contain two almost-identical images, and the subjects had to spot the differences between the two. "Where's Waldo?" images are complex scenes filled with small objects and characters, and the subjects had to find the character Waldo. Each viewing was recorded for 45 seconds.

Eye movements were recorded using an SR Research EyeLink II eye-tracker at 500 Hz sampling frequency. The stimuli were presented such that they are within 36\degree x 25.2\degree of the subjects' visual angle. 480 eye movement samples were obtained from this data set.

\begin{table}[ht]
\centering
\begin{tabular}{ |c|c|c|c|c|c|c| } 
 \hline
 & Hz & Stimuli & Tasks & Subj. & Sample & Time(s)\\ 
 \hline
 \hline
 \multirow{2}{*}{\textbf{EMVIC}} & \multirow{2}{*}{1000} & \multirow{2}{*}{face} & \multirow{2}{*}{free} & \multirow{2}{*}{34} & \multirow{2}{*}{1430} & ave.\\ 
 & & & & & & 2.5s \\
 \hline
 \multirow{2}{*}{\textbf{FIFA}} & \multirow{2}{*}{1000} & \multirow{2}{*}{natural} & free, & \multirow{2}{*}{8} & \multirow{2}{*}{3200} & \multirow{2}{*}{2s}\\ 
 & & & search & & & \\
 \hline
 \multirow{2}{*}{\textbf{ETRA}} & \multirow{2}{*}{500} & natural, & free, & \multirow{2}{*}{8} & \multirow{2}{*}{480} & \multirow{2}{*}{45s}\\ 
 & & puzzle & search & & & \\
 \hline
 \hline
 Total & & & & 50 & 5110 & \\
 \hline
\end{tabular}
\caption{Summary of data sets.}
\label{table:dataset}
\end{table}

\subsection{Data Preprocessing and Augmentation}
To recap, we combine three data sets into a joint data set $D$. Each sample $s \in \mathbb{R}^{(2, t)}$ is a vector with 2 channels (x and y) and a variable length $t$. To work across multiple data sets, we preprocess each $s$ as follows:
\begin{itemize}
    \item We turn blinks (negative values) to zero since not all data sets have blink data.
    \item We standardize to a sampling frequency of 500 Hz. The EMVIC and FIFA data sets are downsampled from 1000 Hz to 500 Hz by dropping every other gaze point.
    \item We modify the coordinates such that the origin (0, 0) is at the top-left corner of the screen. This is to ensure that the network processes eye movements in the same scale.
    \item We scale the coordinates such that a subject's 1\degree of visual angle corresponds to roughly 35 pixels (1 dva $\approx$ 35px). For FIFA and ETRA data sets, these are estimated based on their given eye-tracker and experiment specifications. For EMVIC, we leave the coordinates unprocessed due to lack of details. This is done so that all movements are according to the same visual resolution of the subjects.
\end{itemize}

The inputs to the AEs are standardized into 2-second samples $x \in \mathbb{R}^{(2, t')}$, where $t'$ = 1000 = 500 Hz $\times$ 2s. We increase our data set by taking advantage of the ambiguity of eye movements. For all 5,110 trials in the data sets, we take 2-second time windows that slide forward in time by 20\% or 0.4s, which is equivalent to 200 gaze points. Each window counts as a new sample. With this, the training set size is increased to 68,178 samples.

For all 5,110 trials in the data sets, we take 2s time windows that slide forward in time by 20\% or 0.4s, which is equivalent to 200 gaze points. Using this method, the training set size is increased to 68,178 samples.

\subsection{Velocity Signals}
\begin{figure}
    \centering
    \includegraphics[width=\columnwidth]{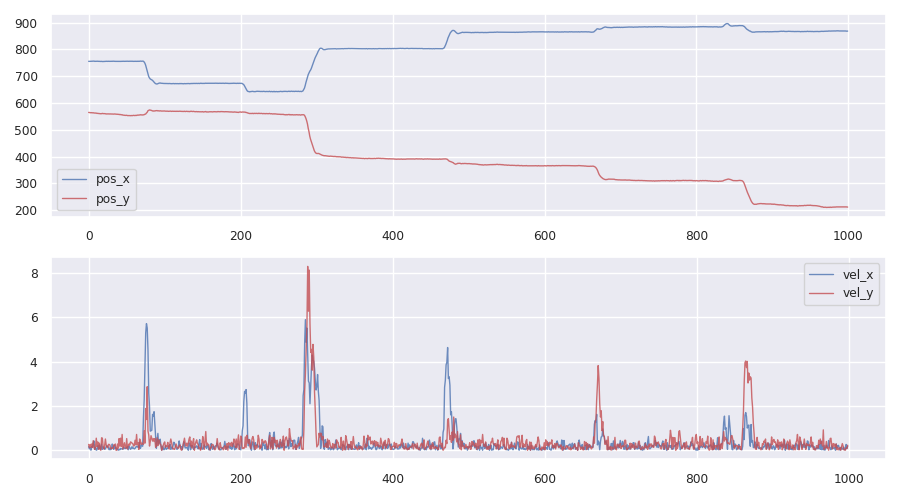}
    \caption{Top: a 2-second position signal at 500 Hz. Bottom: its corresponding velocity signal.}
    \label{fig:pos-to-vel}
\end{figure}

In addition to the raw eye movement data given as a sequence of positions across time (position signals $(x_{pos}, y_{pos})_t$), we also take the derivative, or the rate at which positions change over time (velocity signals $(x_{vel}, y_{vel})_t$), simply calculated as ($\frac{\Delta x}{ms}, \frac{\Delta y}{ms}$). We separately train a position autoencoder (AE\textsubscript{p}) and velocity autoencoder (AE\textsubscript{v}) as they are expected to learn different features. While position signals exhibit spatial information and visual saliency, velocity signals can reveal more behavioral information that may infer a subject's thought process. Velocity is also commonly used as a threshold for eye movement segmentation \cite{eye-tracking-andersson-2017}. Figure \ref{fig:pos-to-vel} shows an example of a position signal and a corresponding velocity signal.
Position signals are further preprocessed by clipping the coordinates to the maximum screen resolution: 1280x1024. For both signals, neither scaling nor mean normalization is done. Based on our experiments, we found that this was especially important for velocity signals.

\subsection{Network Architecture}
\begin{figure}
    \centering
    \includegraphics[width=\columnwidth]{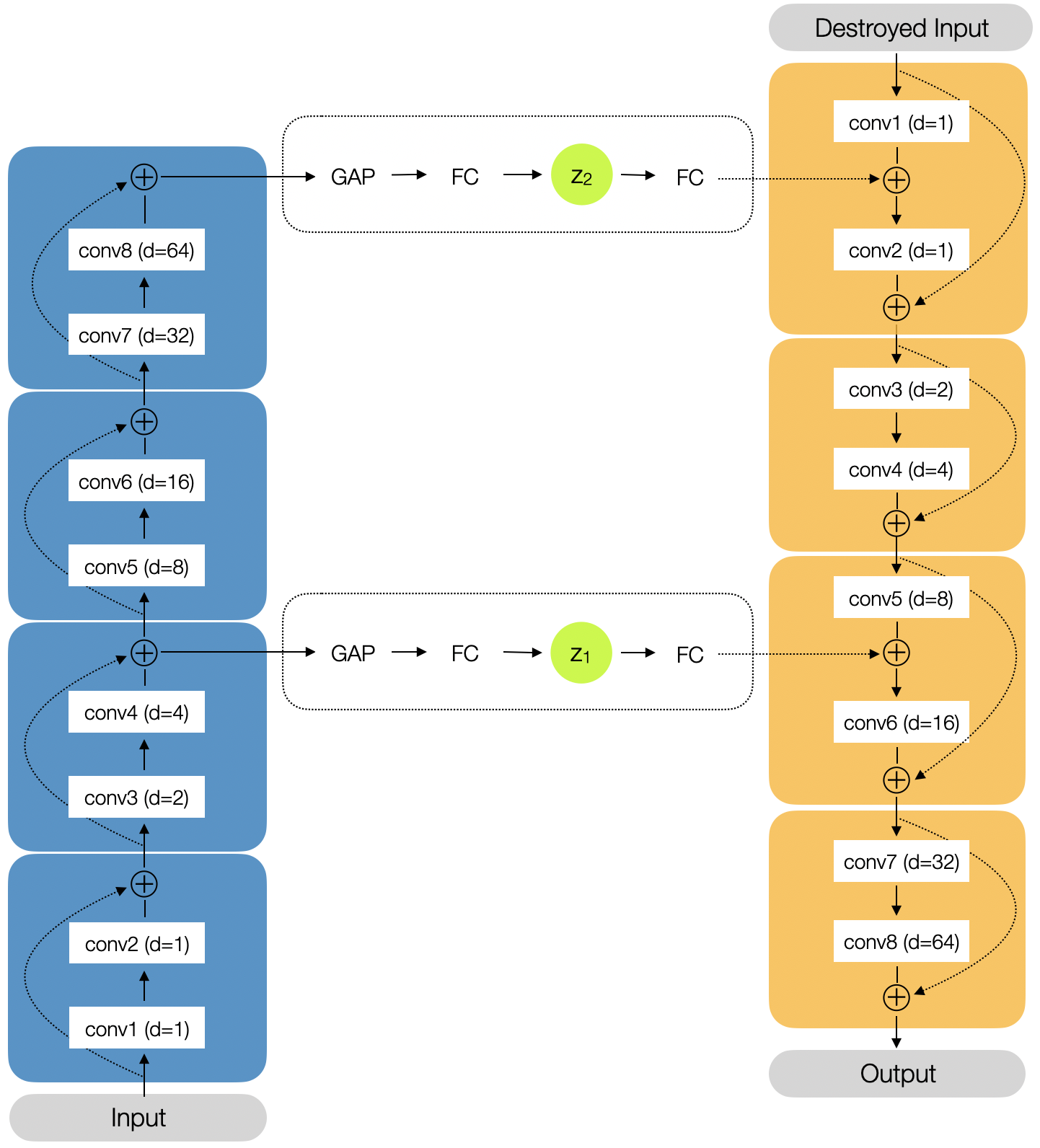}
    \caption{Architecture of the Micro-Macro Autoencoder, with each convolutional layer having a specified dilation.}
    \label{fig:architecture}
\end{figure}

In this subsection, we first describe the TCN architecture of both the encoder and decoder. Next, we describe how a micro and macro representations are learned in the bottlenecks. Lastly, we describe an \emph{interpolative} decoder that \emph{fills in} a destroyed signal to reconstruct or recover the original. The overall architecture of the autoencoder is visualized in Figure \ref{fig:architecture}, and a summary of its main components is shown in Table \ref{tab:network-specs}. The number of filters and layers were chosen empirically.

\begin{table}[h]
    \centering
    \begin{tabular}{|m{10em}|c|c|} 
    \hline
        & \textbf{position AE} (AE\textsubscript{p}) & \textbf{velocity AE} (AE\textsubscript{v}) \\ 
    \hline
    \hline
        Encoder TCN & 128 filters x 8 layers & 256 filters x 8 layers \\
    \hline
        Micro-scale Bottleneck & 64-dim FC & 64-dim FC \\
    \hline
        Macro-scale Bottleneck & 64-dim FC & 64-dim FC \\
    \hline
        \multirow{2}{*}{Decoder TCN} & 128 filters x 4 layers; & \multirow{2}{*}{128 x 8 layers} \\
        & 64 filters x 4 layers & \\
    \hline
        Total Parameters & 652,228 & 1,964,676 \\
    \hline
    \end{tabular}
    \caption{Autoencoder specifications}
    \label{tab:network-specs}
\end{table}

\subsubsection{Convolutional Layers}
The encoder and decoder of the AE are implemented as TCNs. However, the encoder is non-causal in order to take in as much information as possible. The decoder remains causal, as this forces the encoder to learn temporal dependencies.

\begin{figure}
    \centering
    \includegraphics[width=0.8\columnwidth]{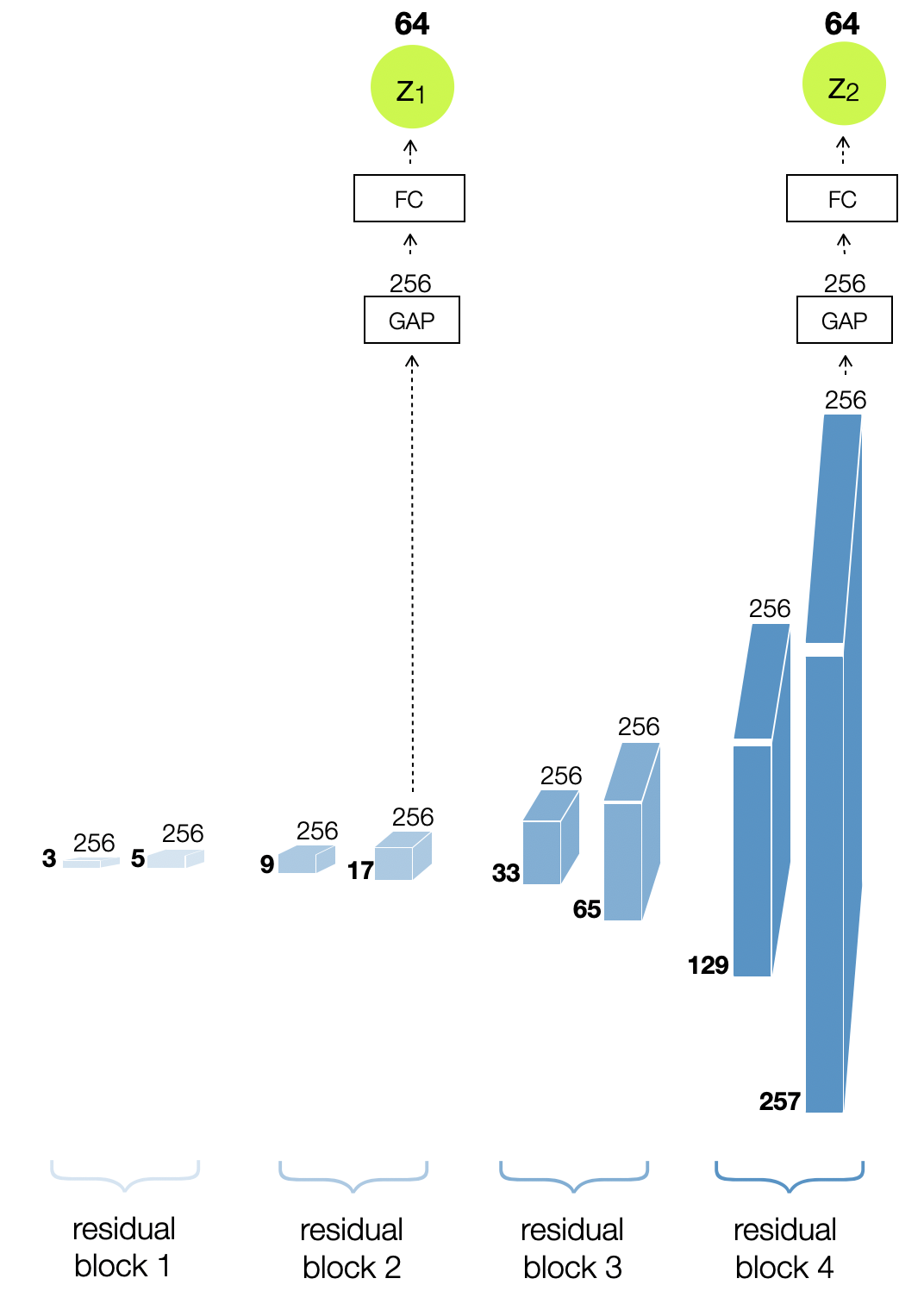}
    \caption{Convolutional layers of AE\textsubscript{v} encoder. The height corresponds to the effective receptive field of each convolution operation. The width corresponds to the number of filters. The outputs at the fourth and eighth layers are calculated to be the micro-scale and macro-scale representations, respectively. Heights roughly to scale.}
    \label{fig:receptive-field}
\end{figure}

Convolutions have a fixed kernel size of 3 and stride 1. Zero-padding is used to maintain the same temporal dimension across all layers. All convolutions are followed by a Rectified Linear Unit (ReLU) activation function and Batch Normalization \cite{batchnorm}. Both the encoder and decoder networks have 8 convolutional layers. These are split into 4 residual blocks \cite{resnet} with 2 convolutional layers each. The layers have exponentially-increasing dilations starting at the second layer (1, 1, 2, 4, 8, 16, 32, 64), resulting in the following receptive fields: (3, 5, 9, 17, 33, 65, 129, 257). Figure \ref{fig:receptive-field} visualizes the growth of the receptive field across layers.

\subsubsection{Bottleneck}
Our AEs have two bottlenecks, each encoding information at different scales. The first takes in the output of the fourth convolutional layer, while the second takes in that of the eighth convolutional layer. Recall that the individual values from these layers were calculated with receptive fields of 17 and 257. Therefore, the first bottleneck can be thought of as encoding micro-scale information, or the fine-grained and fast-changing eye movement patterns. The second encodes macro-scale information, or the flow and slow-changing patterns. This is partly inspired by \cite{deepeye}.

Specifically, the representations at these bottlenecks are learned as follows: first, the convolutional layer outputs are downsampled with a Global Average Pooling (GAP) layer that compresses the time dimension (GAP: $(f, t) \rightarrow (f)$ where $f$ is the number of convolution filters and $t$ is the number of time steps). Then, a fully-connected (FC) layer transforms these downsampled values into micro-scale representation $z_1$ and macro-scale representation $z_2$. The two representations are independent, i.e. there is no forward connection from $z_1$ to $z_2$. From initial experiments, this resulted in better performance. All representations is a feature vector of size 64.

\subsubsection{Interpolative Decoder}
The decoder used in this work is a modification from the vanilla AE architecture. In this model, the original signal is first destroyed by randomly dropping values and then input to the decoder. The task of the decoder remains the same: to output a reconstruction, but it can also now be described as filling in the missing values. Thus, we call it an \emph{interpolative} decoder.

Intuitively, inputting a destroyed version of the original signal to the decoder may free up the encoder to capture more of the nuances in the data, instead of having to also encode the scale and trend of the signal. Representations $z_1$ and $z_2$ act as supplemental information and are used to condition the decoder such that it accurately outputs a reconstruction. $z_2$ is used as an additive bias to the first decoder layer, providing information about the general trend (macro-scale) of the signal. $z_1$ is used as an additive bias to the fifth decoder layer, providing more specific (micro-scale) information and filling in smaller patterns and sequences.

However, reconstructing the input may become a trivial task since too much information is already available to the decoder. In practice, we found that this can be mitigated with a high dropout probability.
The AE\textsubscript{p} uses $p=0.75$, while the AE\textsubscript{v} uses $p=0.66$. Because position signals are much less erratic, a higher dropout probability had to be used to keep the decoder from relying on the destroyed input.
We use this decoder design as an alternative to the more commonly used autoregressive decoders which output one value at a time. We found that the performance was on-par while requiring less training time.

\subsection{Optimization}

To summarize, this work trains a position autoencoder (AE\textsubscript{p}) and a velocity autoencoder (AE\textsubscript{v}) to learn representations $z_p \in \mathbb{R}^{128}$ and $z_v \in \mathbb{R}^{128}$, respectively. Both are concatenations of representations at a micro-scale $z_1 \in \mathbb{R}^{64}$ and a macro-scale $z_2 \in \mathbb{R}^{64}$, i.e. $z_p = [z_{p1}; z_{p2}]$. The training data consists of three data sets combined into a single data set $D$. Each sample $s \in \mathbb{R}^{(2, t)}$ from $D$ is preprocessed into an input vector $x \in \mathbb{R}^{(2, 1000)}$.
For each $x$, an AE is trained to output a reconstruction $\hat{x} \approx x$. The loss function is simply the sum of squared errors (SSE), computed as follows:
\begin{equation}
    SSE = \sum_{t}(x_t - \hat{x_t})^2
\end{equation}

The AEs are trained using Adam \cite{adam} optimizer, with a fixed learning rate of 5e-4. The total number of training samples is 68,178. The batch size for the AE\textsubscript{p} and AE\textsubscript{v} is 256 and 128, respectively. The networks are implemented using PyTorch 1.3.1 \cite{pytorch}, and trained on an NVIDIA GTX 1070 with 8GB of VRAM. Random seeds were kept consistent throughout experiments. AE\textsubscript{p} was trained for 14 epochs (1 epoch $\approx$ 13 mins.) and AE\textsubscript{v} was trained for 25 epochs (1 epoch $\approx$ 38 mins.).

\subsection{Evaluation}

For evaluation, we input the full-length samples and use the AEs to extract representations to be used as input for classification tasks. We evaluate three types of representation: $z_p$ from AE\textsubscript{p}, $z_v$ from AE\textsubscript{v}, and $z_{pv} = [z_p; z_v]$. The classification tasks are the following:

\begin{table}[hbt]
\centering
\begin{tabular}{ |c|c|c|c|c| } 
 \hline
 Classification Task & Data Set & Classes & Samples \\ 
 \hline
 Biometrics & EMVIC & 34 & 837 \\
 Biometrics & all & 50 & 5110 \\ 
 Stimuli (4) & ETRA & 4 & 480 \\
 Stimuli (3) & ETRA & 3 & 360 \\
 Age Group & FIFA & 2 & 3200 \\
 Gender & FIFA & 2 & 3200 \\
 \hline
\end{tabular}
\caption{Classification tasks used for evaluating the representations learned by the autoencoder.}
\label{table:evaluation}
\end{table}

\begin{enumerate}
    \item \emph{Biometrics on EMVIC data set}. We use the official training and test set, reporting accuracies for both. Our results will be compared to the work in \cite{lpitrack}.
    For a fair comparison, we mimic their setup by reporting a 4-fold Cross-Validation (CV) accuracy on the training set, and another on the test set after fitting on the whole training set.

    \item \emph{Biometrics on all data sets}. We combine the three data sets and classify a total of 50 subjects, each with a varying number of samples. In contrast to Biometrics on EMVIC data set, this task is now performed on eye movements from different experiment designs (e.g. eye tracker setup, stimuli used). Therefore, this is a more difficult task and is better suited to evaluate the validity and generalizability of the method.

    \item \emph{Stimuli Classification on ETRA data set}. We use the 4 image types (blank, natural, puzzle, waldo) as labels, where each type has 120 samples. This task, referred to as \emph{Stimuli (4)}, was also done in \cite{encodji, kumar-2019}. Unfortunately, the composition of the data that we use have variations that prohibit us from fairly comparing our work to theirs. Instead, we compare with another work \cite{minhash}, which did the same task but using only 3 labels (natural, puzzle, waldo) with 115 samples each. We use all 120 available samples, but since this is a minor variation from their setup, we still compare our accuracy with theirs. This task, \emph{Stimuli (3)}, is done on a leave-one-out CV (LOOCV) setup to be as similar as possible to theirs.
    \item \emph{Age Group Classification on FIFA data set}. FIFA provides the subjects' ages which range from 18-27. They are split into two groups: 18-22, and 22-27, yielding 1,600 samples per group. A number of previous works have done a similar task, but because they used different data sets, we are not able to fairly compare with their results.
    \item \emph{Gender Classification on FIFA data set}. FIFA was collected from 6 males and 2 females, and we use their gender as labels for their eye movements. The resulting samples are unbalanced, with 2,400 samples for male subjects, and only 800 for females. However, no sampling technique is performed. As with age group classification, there is no previous work with which we can fairly compare with.
\end{enumerate}

To serve as a soft benchmark for tasks without a similar work, we also apply PCA on the position and velocity signals, each with 128 components (PCA\textsubscript{pv}).
The classifier used for all tasks is a Support Vector Machine (SVM) with a linear kernel. Grid search is conducted on the regularization parameter $C = [0.1, 1, 10]$. For all tasks, the accuracy will be reported. Multi-class classification is conducted using a One-vs-Rest (OVR) technique. Unless otherwise stated, all experiments will be conducted in a 5-fold CV setup. PCA, SVM, and CV are implemented using scikit-learn \cite{scikit-learn}.

\section{Results and Discussions}
This section details the classification results and three additional experiments to gauge the representations. For simplicity, we omit the reconstruction errors, as those are not of primary concern when evaluating representations.

\subsection{Classification Tasks}
\subsubsection{Performance}
\label{sec:results}
\begin{table}[htb]
    \centering
    \begin{tabular}{|m{6.5em}|c|c|c|c|c|}
     \hline
     Classification Task & PCA\textsubscript{pv} & $z_p$ & $z_v$ & $z_{pv}$ & others \\
     \hline
     \hline
     Biometrics (EMVIC-Train) & 18.4 & 31.8 & \textbf{\underline{86.8}} & 84.4 & 86.0 \cite{lpitrack}\\ 
     \hline

     \multirow{3}{9em}{Biometrics (EMVIC-Test)} & \multirow{3}{*}{19.7} & \multirow{3}{*}{31.1} & \multirow{3}{2em}{\textbf{\underline{87.8}}} & \multirow{3}{*}{\underline{87.8}} & 81.5 \cite{lpitrack}\\
     & & & & & 82.3*\\
     & & & & & 86.4*\\
     \hline
     Biometrics (All) & 24.6 & 29.0 & \underline{79.8} & 78.4 & - \\
     \hline

     Stimuli (4) & {38.8} & 81.3 & 85.4 & \underline{87.5} & - \\
    \hline

     Stimuli (3) & 55.8 & 90.3 & 87.2 & \textbf{\underline{93.9}} & 88.0** \cite{minhash} \\
     \hline

     Age Group & 62.0 & 61.9 & \underline{77.7} & 77.3 & - \\

     \hline
     Gender & 51.12 & 54.9 & 85.8 & \underline{86.3} & - \\
     \hline
    \end{tabular}
    \caption{Accuracies for various classification tasks. Underlined numbers are highest among AE models; bold numbers are highest among different works.
    \newline
    * \small{these were mentioned in \cite{lpitrack} but no citation was found.}
    \newline
    ** \small{their classification used 115 samples per label, ours used 120.}}
    \label{tab:classification-acc}
\end{table}

The results of the classification tasks, along with chance accuracies and other works are summarized in Table \ref{tab:classification-acc}.
First, it is clear that velocity representations $z_v$ carry more discriminative information than $z_p$, as it can perform well on its own and can be supplementary to $z_p$ as in the case of stimuli and gender classification. The performance of $z_p$ only came close to $z_v$ in the stimuli classification task, which is expected since spatial information is explicitly linked to the stimuli.
Next, AE performance on Biometrics task on EMVIC data set was able to outperform the work in \cite{lpitrack}. They used a statistical method to extract spatial, temporal, and static shape features, on which they fitted a logistic regression classifier. They additionally mentioned two works which achieved higher test accuracies (82.3\% and 86.4\%) than theirs, but those were uncited and no document describing those works have been found as of writing. Nevertheless, AE\textsubscript{v} also outperforms those two works.

On stimuli classification on ETRA data set, our work outperformed \cite{minhash}. Recall, however, that the comparison is not entirely balanced due to different number of samples.
The four other tasks have no other works to directly compare with, however, we found the performance more than satisfactory. $z_v$ and $z_{pv}$ performed well on the Biometrics task on all data sets despite the fact that the eye movements were gathered from a diverse set of images. This may indicate that the speed and behavior of eye movements are sufficient identity markers, and future eye movements-based Biometric systems need not meticulously curate the stimuli used for interfaces.
For age group and gender classification, note that the task is performed only with 8 subjects. In terms of viability of eye movements for classifying a person's demographic, these results are inconclusive. Nevertheless, the accuracies are well above chance and PCA feature extraction, encouraging further experimentation on the area.

\subsubsection{Feature Analysis}
To further inspect the importance of the representations, we take the linear SVM fitted on $z_{pv}$ (total 256 total dimensions), and inspect the top 20\% features. Though the linear SVM may suffer from fitting on a large number of dimensions, this presents an estimate of how useful the feature types are for various tasks.

\begin{figure}[!hbt]
    \centering
    \subfloat{
        {\includegraphics[width=\columnwidth]{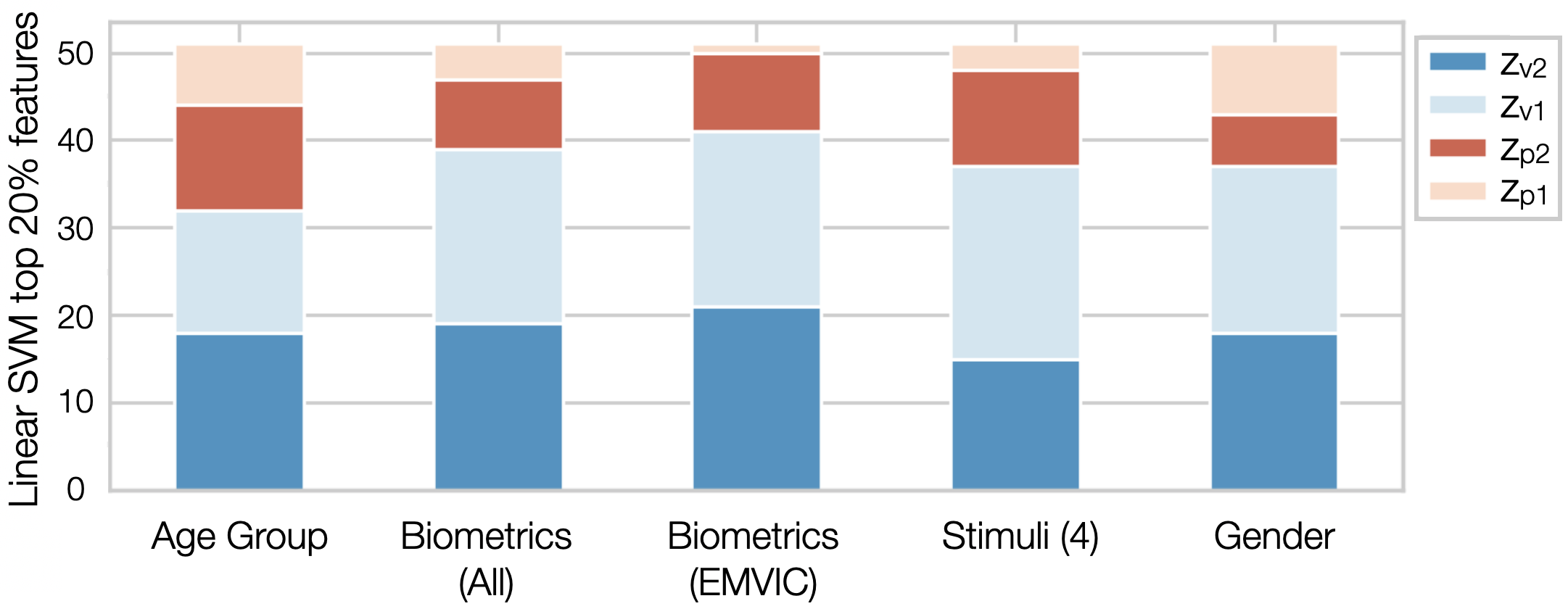} }}%
    \caption{Count of features per type among the top 20\% weights of the trained linear SVM.}
    \label{fig:feature-importances}
\end{figure}
Figure \ref{fig:feature-importances} shows the result. Velocity representations dominate the top features. Both the micro and macro scales of the velocity signal are useful, though the micro-scale takes a slightly larger share of the top features. Position representations are much less important, even on the stimuli classification task. Thus, a velocity autoencoder may be a less complicated but sufficient method for representing eye movements. However, this may still be explored with other classification tasks.

\begin{figure}[hbt]
    \centering
    \includegraphics[width=\columnwidth]{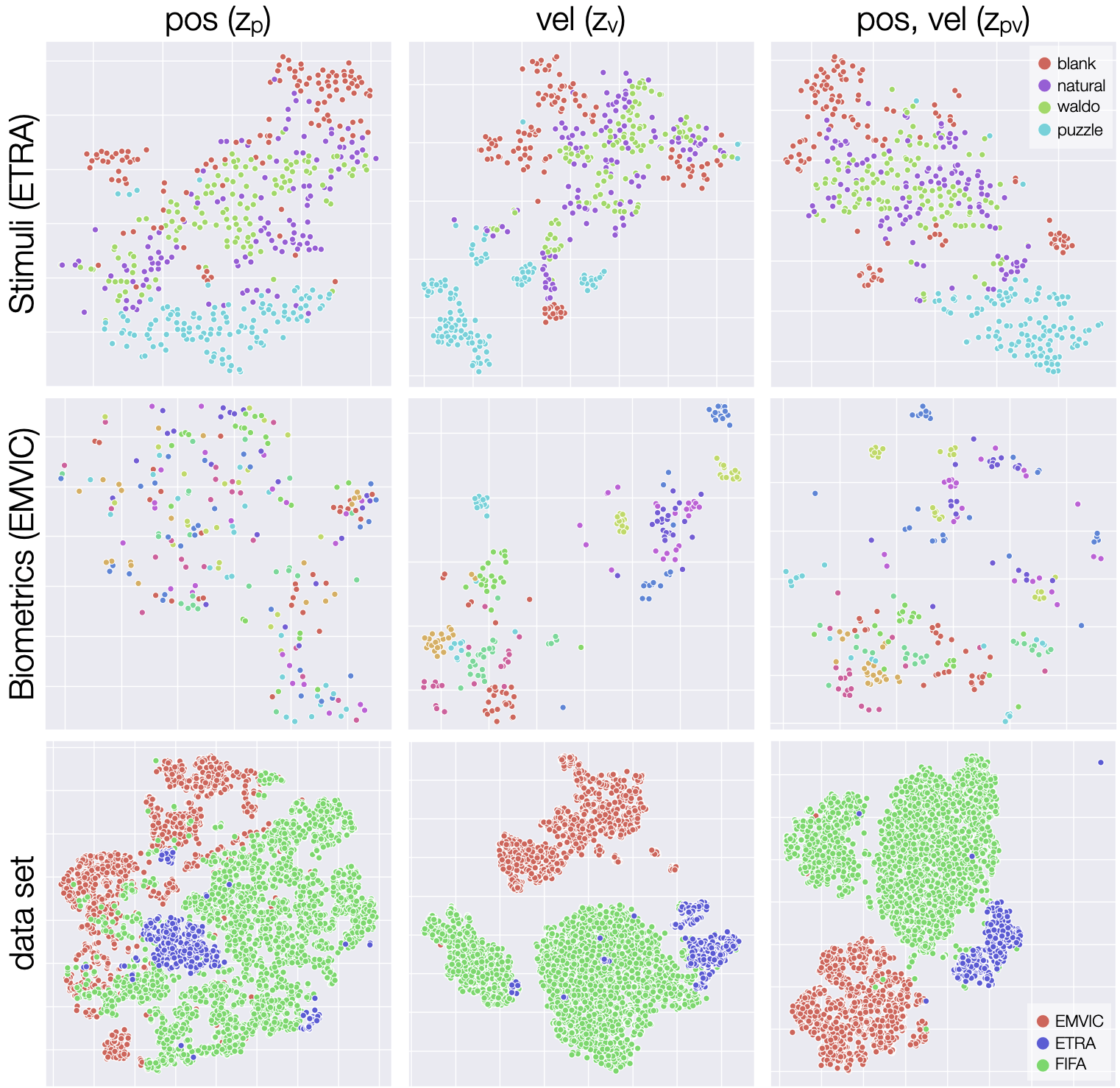}
    \caption{t-SNE visualizations of learned representations. Top: ETRA samples, labels are stimuli types. Middle: EMVIC samples, labels are 10 subjects with most number of samples. Bottom: all samples, labels are the data sets.}
    \label{fig:tsne-embedding}
\end{figure}

Next, we explore the representations by visualizing the embedding space. We apply t-SNE \cite{tsne}, a dimensionality reduction algorithm that preserves the distances of all points, on $z_p$, $z_v$, and $z_{pv}$, as shown in Figure \ref{fig:tsne-embedding}. Consistent with the accuracies in Table \ref{tab:classification-acc}, $z_p$ and $z_v$ are able to discriminate stimuli types. Visualization of $z_p$ on Biometrics show almost no clustering, while $z_v$ exhibits some. We also plot all samples and label them according to their data sets. Clear clustering can be observed based on $z_v$. This is made clearer when $z_p$ and $z_v$ was combined, showing that these two representations can be indeed supplementary.

\subsection{Additional Experiments}
\subsubsection{Generalizability}
To test if the AE is generalizable and did not overfit, we use AE\textsubscript{v} to extract representations for unseen samples. The Biometrics task is performed using the data set provided in \cite{mit-lowres}, herein termed as MIT-LowRes. This data set contains eye movement signals from 64 subjects looking at 168 images of varying low resolutions. Only the samples obtained from viewing the highest-resolution will be used for this experiment. This corresponds to 21 samples for 64 subjects, amounting to 1,344 total samples. The data was recorded in 240 Hz. To be used for the AE\textsubscript{v} model, the signals are upsampled to 500Hz using cubic interpolation.

We also train two more AE models. One is trained using the three original data sets but on a 250Hz sampling frequency (AE\textsubscript{v}-250), and the second is trained exclusively on MIT-LowRes at 250Hz (AE\textsubscript{v}-MLR). These models use the same architecture and specifications as AE\textsubscript{v}, and we only modify the dilations so that the receptive field is approximately halved.

\begin{table}[h]
    \centering
    \begin{tabular}{|m{11em}|c|c|c|}
        \hline
         Classification Task & AE\textsubscript{v} & AE\textsubscript{v}-250 & AE\textsubscript{v}-MLR \\
         \hline
         \hline
         Biometrics (MIT-LowRes) & \underline{23.7} & 21.5 & 18.38 \\
         \hline
    \end{tabular}
    \caption{Accuracies for a Biometrics task on MIT-LowRes, an unseen data set. For comparison, AE\textsubscript{v}-MLR is a model trained exclusively on MIT-LowRes.}
    \label{tab:unseen-results}
\end{table}
From Table \ref{tab:unseen-results}, we see that
AE\textsubscript{v} achieved the highest accuracy of the three models. It outperformed AE\textsubscript{v}-250, showing that there are indeed more meaningful information with a higher sampling frequency. However, even AE\textsubscript{v}-250 outperformed AE\textsubscript{v}-MLR. This shows that the AEs benefited from training on more data, and can indeed generalize to unseen samples, even if they're from another data set.  Furthermore, this also shows that signals at 240Hz can be upsampled to 500Hz through simple cubic interpolation in order to benefit from 500Hz models.

\subsubsection{Input Length / Viewing Time}
The use of a GAP layer enables the autoencoder to take in inputs of any length. Recall that we train the AEs on only 2s, and we evaluated it with the full-length samples. In this experiment, we explicitly test for the effect of the input length or viewing time on the representations. We do this by using 1s, 2s, averaged representations of disjoint 2-second segments (2s*), and full-length inputs to AE\textsubscript{v}. From Table \ref{tab:viewing-time}, it can be seen that the AE can scale well even up to 45s without loss of performance, making it more usable on any eye movement sample.

\begin{table}[htb]
    \centering
    \begin{tabular}{|m{11em}|c|c|c|c|}
     \hline
     Classification Task & 1s & 2s & 2s* & full\\
     \hline
     \hline
     Biometrics (EMVIC-Train) & 78.9 & 84.2 & 83.35 & \underline{86.8} (22s) \\ 
     \hline
     Biometrics (EMVIC-Test) & 79.0 & 85.6 & 86.6 & \underline{87.8} (22s) \\ 
     \hline
     Biometrics (All) & 69.3 & 76.9 & 79.7 & \underline{79.8} (45s) \\
     \hline
     Stimuli (4) & 46.7 & 59.2 & 85.0 & \underline{85.4} (45s) \\
     \hline
     Age Group & 75.1 & 78.2 & - & - \\
     \hline
     Gender & 79.4 & 85.9 & - & - \\
     \hline
    \end{tabular}
    \caption{Accuracies for classification tasks depending on the viewing time (length of input to the AE). 1s = 500 gaze points = 500 time steps}
    \label{tab:viewing-time}
\end{table}

\subsubsection{Comparison with Supervised TCN}
\begin{table}[htb]
    \centering
    \begin{tabular}{|m{11em}|c|c|}
     \hline
     Classification Task & AE\textsubscript{v} (unsupervised) & TCN\textsubscript{v} (supervised) \\
     \hline
     \hline
     Biometrics (EMVIC-Train) & 86.8 & 93.6 \\
     \hline
     Biometrics (EMVIC-Test) & 87.8 & 95.5 \\
     \hline
     Biometrics (All) & 79.8 & 84.5 \\
     \hline
     Stimuli (4) & 89.2 & 90.0 \\
     \hline
     Age Group & 78.0 & 96.8 \\
     \hline
     Gender & 87.4 & 96.2 \\
     \hline
    \end{tabular}
    \caption{Accuracies for classification tasks of AE\textsubscript{v} compared to TCN\textsubscript{v}, supervised version of the encoder network.}
    \label{tab:vs-supervised}
\end{table}

Finally, AE\textsubscript{v} is compared against a supervised TCN (TCN\textsubscript{v}) with the same architecture as the encoder in AE\textsubscript{v}. To be supervised, we add an FC and Softmax layer to the network to output class probabilities.

For each task, we train a new TCN\textsubscript{v} for 100 epochs with early stopping. We perform 4-fold CV for Biometrics (EMVIC), and 5-fold on all other tasks. Table \ref{tab:vs-supervised} shows the results. TCN\textsubscript{v} models clearly outperform AE\textsubscript{v} which is an expected result given that supervised networks tune their weights according to the task. It is, however, encouraging to find that AE\textsubscript{v} can reach as low as 0.8\% difference in accuracy when compared to TCN\textsubscript{v}. AEs also have less tendency to overfit and can be reused for different scenarios.
\section{Related Work}
Our work aims to learn generalizable representations for eye movements through unsupervised learning. To the best of our knowledge, no work with the exact same goal has been done. Related but tangential works that construct gaze embeddings include \cite{cv-gaze-embedding} and \cite{ semantic-gaze-embedding}. The first used eye movement parameters, grids, and heatmaps, while the second used a CNN to extract feature vectors at fixated image patches. Another related work is \cite{encodji} which used a generative adversarial network (GAN) to represent scanpaths. However, theirs is only a small-scale experiment primarily focused on scanpath classification.
\section{Conclusion}
In this work, we proposed an autoencoder (AE) that learns micro and macro-scale representations for eye movements. We trained a position AE and a velocity AE using three different data sets, and we evaluate the representations with various classification tasks. We were able to achieve competitive results, outperforming other works despite using an unsupervised feature extractor and fitting with only a linear classifier. Further experiments showed that the proposed AE can handle any input length, and is able to generalize to unseen samples from a different data set. Performance was also shown to be comparable to a supervised version of the encoder CNN. This work is therefore a positive step towards adapting eye tracking technology to real-world tasks.

\bibliographystyle{IEEEtran}
\bibliography{IEEEabrv,biblio}
%



\end{document}